\documentclass[]{bytedance_seed}

\usepackage[toc,page,header]{appendix}
\usepackage{etoolbox}

\usepackage{natbib}

\usepackage{CJKutf8}

\usepackage{xargs}  

\usepackage{todonotes}  

\usepackage{multirow}

\usepackage{amsmath}
\usepackage{dsfont}

\usepackage{subcaption}

\usepackage{svg}

\usepackage{float}
\usepackage{algpseudocode}
\usepackage[ruled,vlined]{algorithm2e}
\usepackage{adjustbox}
\usepackage{multirow}
\usepackage{multicol}
\usepackage{tcolorbox}
\usepackage{xcolor}
\usepackage{threeparttable,booktabs,multirow,xcolor}
\usepackage[table]{xcolor}
\usepackage{amsmath,amssymb}

\usepackage{graphicx}
\usepackage{subcaption}
\usepackage{cleveref}

\definecolor{catgray}{gray}{0.9}
\definecolor{skyblue}{rgb}{0.53,0.81,0.92} 

\colorlet{skyblue!30}{skyblue!30!white} 

\definecolor{customblue}{RGB}{70,130,180}  
\usepackage[table]{xcolor}  
\usepackage{colortbl}       
\usepackage{tabularx} 
\usepackage{makecell} 
\usepackage[dvipsnames]{xcolor}

\newcolumntype{g}{>{\columncolor{gray!10}}c} 

\definecolor{catgray}{gray}{0.9}
\definecolor{skyblue}{rgb}{0.53,0.81,0.92} 

\colorlet{skyblue!30}{skyblue!30!white} 

\definecolor{customblue}{RGB}{70,130,180}  

\newtcolorbox{evolbox}[2][]{%
  enhanced,
  colframe=customblue,
  colback=white,
  coltitle=white,
  rounded corners,
  boxrule=1pt,
  titlerule=0pt,
  toptitle=1mm,
  bottomtitle=1mm,
  fonttitle=\bfseries,
  width=#2\textwidth, 
  #1
}

\usepackage{minitoc}
\usepackage{url}

\title{%
  \raisebox{-0.2\height}{\includegraphics[height=1.5em]{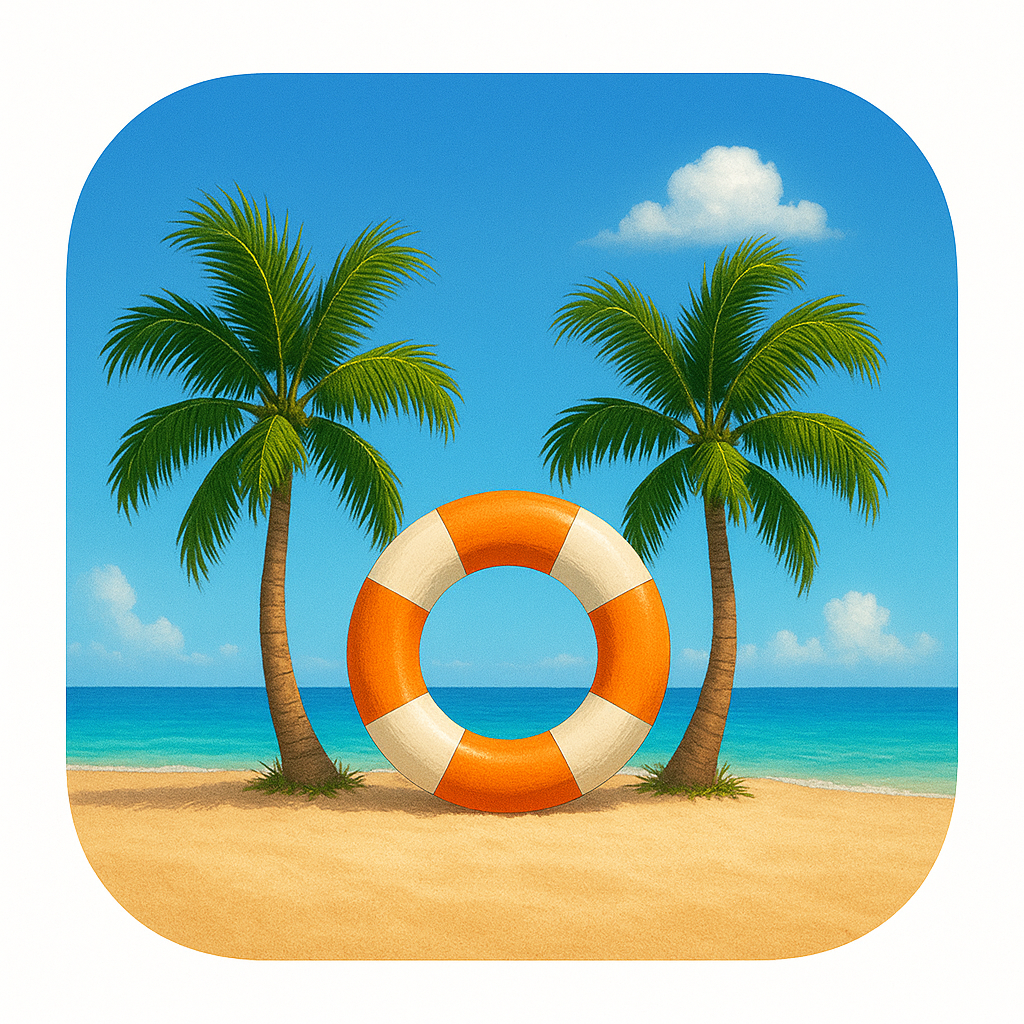}}%
  \hspace{0.2em}%
  LoL: Longer than Longer, Scaling Video Generation to Hour%
}

\author{
  Justin Cui\texorpdfstring{$^{1,2}$}{} \hspace{0.1cm}
  Jie Wu\texorpdfstring{$^{2,\dagger}$}{} \hspace{0.1cm}
  Ming Li\texorpdfstring{$^{2,3}$}{} \hspace{0.1cm}
  Tao Yang\texorpdfstring{$^{2}$}{} \hspace{0.1cm}
  Xiaojie Li\texorpdfstring{$^{2}$}{} \hspace{0.1cm}
  Rui Wang\texorpdfstring{$^{2}$}{} \\ \hspace{0.1cm} 
  Andrew Bai\texorpdfstring{$^{1}$}{} \hspace{0.1cm}
  Yuanhao Ban\texorpdfstring{$^{1}$}{} \hspace{0.1cm}
  Cho-Jui Hsieh\texorpdfstring{$^{1,\ddagger}$}{}
}

\affiliation[1]{UCLA}
\affiliation[2]{ByteDance Seed}
\affiliation[3]{University of Central Florida}
\contribution[\ddagger]{Corresponding author} 
\contribution[\dagger]{Project lead}

\begin{document}

\abstract{
Recent research in long-form video generation has shifted from bidirectional to autoregressive models, yet these methods commonly suffer from error accumulation and a loss of long-term coherence. While attention sink frames have been introduced to mitigate this performance decay, they often induce a critical failure mode we term sink-collapse: the generated content repeatedly reverts to the sink frame, resulting in abrupt scene resets and cyclic motion patterns. Our analysis reveals that sink-collapse originates from an inherent conflict between the periodic structure of Rotary Position Embedding (RoPE) and the multi-head attention mechanisms prevalent in current generative models. To address it, we propose a lightweight, training-free approach that effectively suppresses this behavior by introducing multi-head RoPE jitter that breaks inter-head attention homogenization and mitigates long-horizon collapse. Extensive experiments show that our method successfully alleviates sink-collapse while preserving generation quality. To the best of our knowledge, this work achieves the first demonstration of real-time, streaming, and infinite-length video generation with little quality decay. As an illustration of this robustness, we generate continuous videos up to 12 hours in length, which, to our knowledge, is among the longest publicly demonstrated results in streaming video generation.
}

\maketitle



\section{Introduction}
\label{sec:intro}
\begin{figure}[t]
\centering
\includegraphics[width=0.95\textwidth]{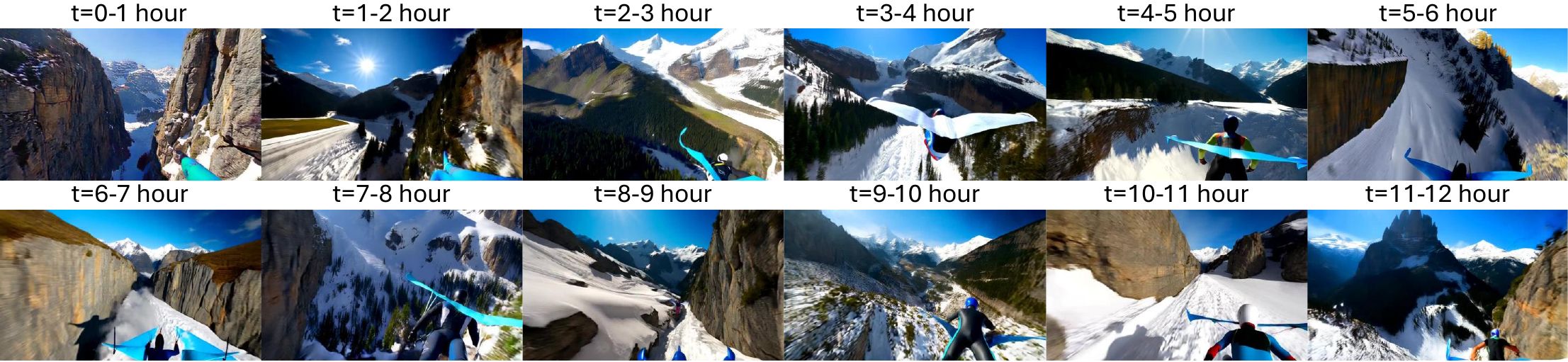}
\caption{Streamingly generated ultra long video (12 hours) for prompt ``A cinematic third-person shot of a wingsuit flyer racing through a narrow mountain valley. The flyer dives downwards, weaving smoothly between jagged cliffs as snow-capped peaks..''.}
\label{fig:teaser}
\end{figure}
Video generation has undergone rapid evolution, propelled by advances in diffusion-based generative modeling. Breakthrough systems such as Sora~\citep{openai2024sora}, CogVideoX~\cite{yang2024cogvideox}, Wan~\citep{wan2025}, Hunyuan-DiT~\citep{kong2024hunyuanvideo} and Veo~\citep{deepmind_veo} have  pushed the frontier of realism in synthesized motion and visual fidelity. These models exhibit remarkable capability in capturing complex spatiotemporal dynamics, producing results that often blur the boundary between generated and real footage. However, they all suffer from heavy computation cost when extending to long sequences. 

To address this limitation, recent work shifts from bidirectional designs to autoregressive generation, predicting each new frame from previously generated ones to support much longer temporal modeling. Among them, Diffusion-Forcing~\citep{chen2024diffusion,kim2024fifo} applies frame-wise noise schedules but often suffers from instability~\citep{chen2025skyreels,he2025matrix}. Simpler approaches predict the next frame from clean contexts using KV cache for efficient streaming. CausVid~\citep{yin2025causvid} distills a bidirectional teacher into a streaming student but faces overexposure from overlapping frames and train–test mismatch. Self Forcing~\citep{huang2025self} alleviates this via distribution alignment, yet remains constrained by the teacher’s temporal limit. Recent works such as LongLive~\citep{yang2025longlive}, Self-Forcing++~\citep{cui2025self}, and Rolling-Forcing~\citep{liu2025rollingforcingautoregressivelong} extend high-quality generation to minutes through windowed DMD on ultra-long sequences.

Using attention sink to stabilize  autoregressive models  was first proposed in StreamingLLM~\citep{streamingllm} which showed that retaining KV pairs of initial tokens can recover performance lost in windowed attention in Large Language Models. This concept has recently been adopted in autoregressive video generation~\citep{yang2025longlive,shin2025motionstreamrealtimevideogeneration,huang2025self} to enhance alignment and stability.
Despite differing training paradigms, state-of-the-art methods like LongLive~\citep{yang2025longlive} and Self-Forcing++~\citep{cui2025self} exhibit a shared failure when using attention sink: frames repeatedly regress toward the sink frames, a phenomenon we term sink-collapse. For example, both methods collapse at exact latent frame index 132 and 201 as shown in~\cref{fig:main_figure}, with more collapses emerging regardless of input noise or prompts. Repetition has also been observed when extending bidirectional models as shown in RIFLEx~\citep{zhao2025riflex}, however it attributes the repetition to a single temporal dimension which does not generalize to autoregressive settings.

In this paper, we introduce Longer than Longer (LoL), the first attempt to tackle the sink-collapse problem in state-of-the-art autoregressive video generation models and propose a way to extend streaming generation indefinitely.
First, by examining the  repetitive patterns in Self-Forcing++~\citep{cui2025self} and LongLive~\citep{yang2025longlive}, we find that although both exhibit sink-collapse at identical frame indices, no clear periodicity emerges, unlike the periodic behavior observed in bidirectional models such as RIFLEx~\citep{zhao2025riflex}. Summing phase alignment around sink frames from all temporal dimensions reveals that collapse points coincide with local maxima, indicating that sink-collapse arises from multiple temporal dimensions.
Second, since modern transformers rely on multi-head attention across subspaces, we analyze their attention patterns and find that collapse occurs when multiple heads simultaneously assign significantly high weights to sink frames. Motivated by these observations, we propose a simple yet effective remedy by shifting the base frequencies of different heads around the original base $\theta$ which reduces inter-head homogenization and mitigates collapse.
Finally, we extend video length from multiple minutes~\cite{yang2025longlive,cui2025self} to infinity by integrating streaming RoPE generation and noise sampling, and a 3D causal VAE decoder at inference, enabling continuous video generation with sustained quality.
In summary, our contributions are:

\begin{itemize}[leftmargin=*]
\item  We systematically analyze the phenomenon of sink collapse in state-of-the-art ultral long video generation models and reveal the root cause of such phenomenon in autoregressive long video generation.

\item We propose a simple yet highly effective method to mitigate sink collapse by shifting the frequencies of different attention heads. Together with our implementation of streaming RoPE generation and noise sampling, we are able to extend real-time video generation to indefinitely long without collapsing.

\item Extensive experiments show that our method preserves generation quality while effectively mitigating the sink-collapse phenomenon compared to baseline methods.
\end{itemize}

To the best of our knowledge, this is the first time that real-time infinite streaming generation is achieved with little quality degradation, while solely relying on models of 1.3B size and KV cache.
\section{Related Work}
\label{sec:related_works}
\begin{figure*}[t]
    \centering
    \includegraphics[width=1.0\linewidth]{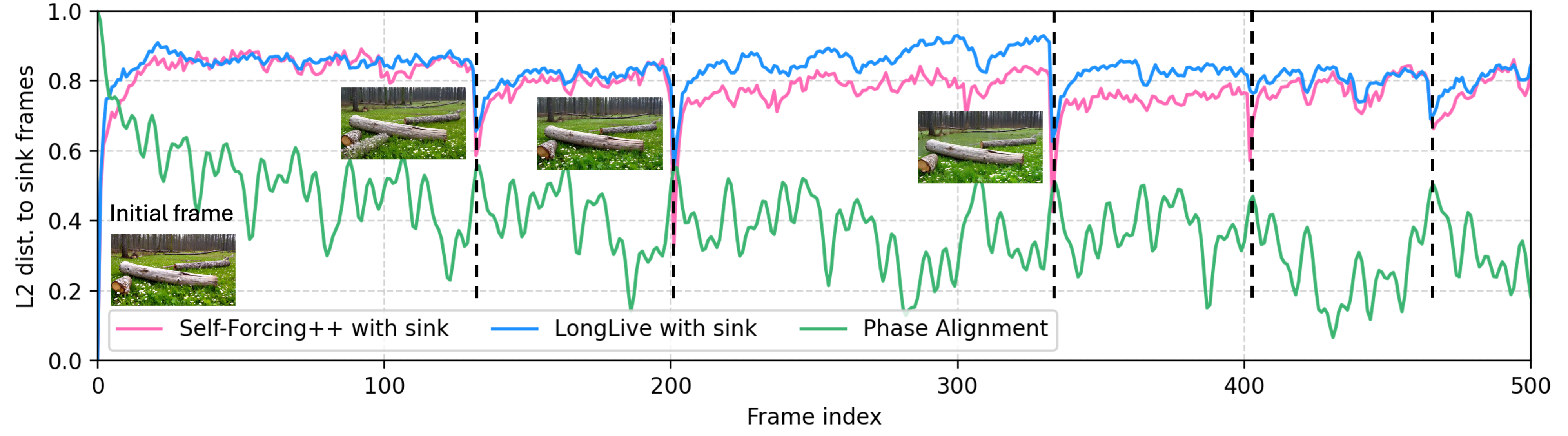}
    \vspace{-0.7cm}
    \caption{The plot of intra-head phase concentration which shows the normalized L2 distance to the sink frames against the mean phase of RoPE embeddings (base 10000) with respect to the sink frames. The results reveal that sink-collapse emerges almost exactly where the phase concentration attains local maxima. Besides sink-collapse events, additional drops in L2 distance also occur around local maxima.}
    \label{fig:sink-collapse}
    \vspace{-0.4cm}
\end{figure*}

\textbf{Long Video Generation}
To extend the generation to long sequences, a number of techniques have been introduced~\citep{jiang2025lovic,kodaira2025streamdit,fang2025inflvg,lin2025autoregressive,deng2024autoregressive,jin2024pyramidal,zhang2025pretraining,hong2025relic,lu2025reward,guo2025end}. SkyReels-V2~\cite{chen2025skyreels} and MAGI-1~\cite{teng2025magi} adopt diffusion-forcing~\citep{chen2024diffusion} to support potentially infinite rollouts. CausVid~\citep{yin2025causvid} employs asymmetric distillation and block causal attention in training and a KV cache at inference time to autoregressively extend sequences. Self-Forcing~\citep{huang2025self} further aligns training with inference by incorporating the KV cache directly during training, producing high-quality short videos. FIFO-Diffusion~\cite{kim2024fifo} utilizes   a first-in-first-out queue to perform heterogeneous denoising using a regular bi-directional diffusion model. APT~\cite{lin2025autoregressive} distills the model into 1-step generator and utilizes adversarial training with KV cache to extend the generation up to 60 seconds.

\textbf{Ultra Long Video Generation}
Recently, the landscape of ultra-long video generation has advanced rapidly, marked by several key innovative methods such as Rolling-Forcing~\cite{liu2025rollingforcingautoregressivelong}, which establishes a diffusion-forcing training paradigm; LongLive~\cite{yang2025longlive}, which employs attention sinks for temporal coherence and KV-recaching for seamless prompt transitions; and Self-Forcing++~\cite{cui2025self}, which extends the DMD formulation to long-horizon modeling, significantly increasing the scalability of video diffusion models. SVI~\cite{li2025stable} utilizes a hybrid approach to generate motion and content-rich long videos. For the first time, these approaches enable the synthesis of multi-minute videos while preserving high visual fidelity and temporal stability. 

\textbf{Position Embedding}
To capture sequential dependencies among tokens, positional embeddings encode order information into input representations. Early approaches such as sinusoidal embeddings~\cite{vaswani2017attention} assign fixed position vectors to each token, while later methods introduce learnable embeddings to adapt positions during training~\cite{devlin2019bert,dosovitskiy2020image}. More advanced designs, including relative position encodings~\citep{shaw2018self} and Rotary Position Embedding~\citep{su2024roformer}, model positional relationships through relative distances or rotations in feature space, improving generalization to longer sequences. Extensions like YaRN~\citep{peng2023yarn} further enhance extrapolation by rescaling frequency parameters. Unlike these studies that focus on interpolation and extrapolation, our work instead investigates a distinct issue, namely the sink-collapse phenomenon in autoregressive video generation~\citep{cui2025self,yang2025longlive,yesiltepe2025infinity}.
\section{Method}
\label{sec:method}
In this section, we first introduce the background for transforming bi-directional models into autoregressive ones for ultra-long video generation, then identify and address the cause of sink-collapse. We then show how to achieve infinite streaming generation under the current setting.

\subsection{Background}
\textbf{Autoregressive Long Video Generation}
Due to high computational cost, state-of-the-art bidirectional models~\cite{deepmind_veo,wan2025,li2025hunyuan,gao2025seedance} are limited to generating only a few seconds of video. Recently, LongLive~\cite{yang2025longlive} and Self-Forcing++~\cite{cui2025self} employ a short-horizon bidirectional teacher to extend generation to several minutes, utilizing asymmetric distillation based on CausVid~\cite{yin2025causvid} and Self-Forcing~\cite{huang2025self}, reaching nearly 99\% of the base model’s positional embedding capacity. The core methodology behind these methods extends regular Distribution Matching Distillation~\cite{yin2024improved} which computes the reverse KL loss $\mathbb{E}_t \!\left[ D_{\mathrm{KL}}\!\left(p_t^{\mathrm{gen}} \,\Vert\, p_t^{\mathrm{data}}\right) \right]$ on windowed self-generated long sequences with backward noise initialization~\cite{yang2025longlive,cui2025self}, which can be formulated as
\begin{equation}
\begin{aligned}
\nabla_{\theta}\mathcal{L}_{\substack{\text{DMD} \\ \text{extended}}}
& \approx
-\,\mathbb{E}_{t}\,\mathbb{E}_{\,z_i}
\Biggl[\int\!\Bigl(
         s^{T}\!\bigl(\Phi(G_{\theta}(z_i),t),t\bigr)
         - s^{S}\!\bigl(\Phi(G_{\theta}(z_i),t),t\bigr)
      \Bigr)\,
      \frac{dG_{\theta}(z_i)}{d\theta}\,dz_i
   \Biggr].
\end{aligned}
\label{eq:dmd_extended}
\end{equation}

\textbf{Rotary Position Embedding (RoPE)}
To enable the model to differentiate tokens across positions, positional embeddings are introduced to inject positional information into input representations~\citep{vaswani2017attention, su2024roformer, bloc97, peng2023yarn}.
Among these approaches, Rotary Position Embedding (RoPE)~\citep{su2024roformer} has become one of the most widely used methods in transformer-based architectures.
RoPE encodes temporal or spatial positions by rotating the query and key vectors within a complex plane according to their token indices, effectively allowing the model to capture relative positional relationships. For a given position index \( t \) and channel dimension \( i \), the rotation is defined as
\begin{equation}
\theta_i = \theta^{-\frac{2i}{d}}, \qquad
\mathbf{R}(t, i) =
\begin{pmatrix}
\cos(t\,\theta_i) & -\sin(t\,\theta_i) \\
\sin(t\,\theta_i) &  \cos(t\,\theta_i)
\end{pmatrix},
\end{equation}
where \( d \) is the hidden dimension and \( \theta \) (typically \( 10{,}000 \)) controls the base frequency. The rotated query and key vectors are obtained by
\begin{equation}
\mathbf{q}'_m = \mathbf{R}(m, i)\,\mathbf{q}_m, \qquad
\mathbf{k}'_n = \mathbf{R}(n, i)\,\mathbf{k}_n.
\end{equation}
When computing attention between positions \( m \) and \( n \), their rotated forms yield
\begin{equation}
\langle \mathbf{q}'_m, \mathbf{k}'_n \rangle 
= \langle \mathbf{q}_m,\, \mathbf{R}(n - m)\,\mathbf{k}_n \rangle,
\label{eq:rope_relative}
\end{equation}
where \( \mathbf{R}(n - m) \) represents a rotation parameterized by the relative offset \( n - m \) indicating that the attention score depends on positional differences, enabling RoPE to naturally encode relative positional information.

\subsection{Sink Frames \& Sink Collapse}
Attention sink is first proposed in StreamingLLM~\cite{streamingllm} to extend LLM generation to significantly long sequences. Recent work such as LongLive~\cite{yang2025longlive,liu2025rollingforcingautoregressivelong} extends it to autoregressive ultra long video generation for more stable generations.

\begin{figure*}
    \centering
    \includegraphics[width=\linewidth]{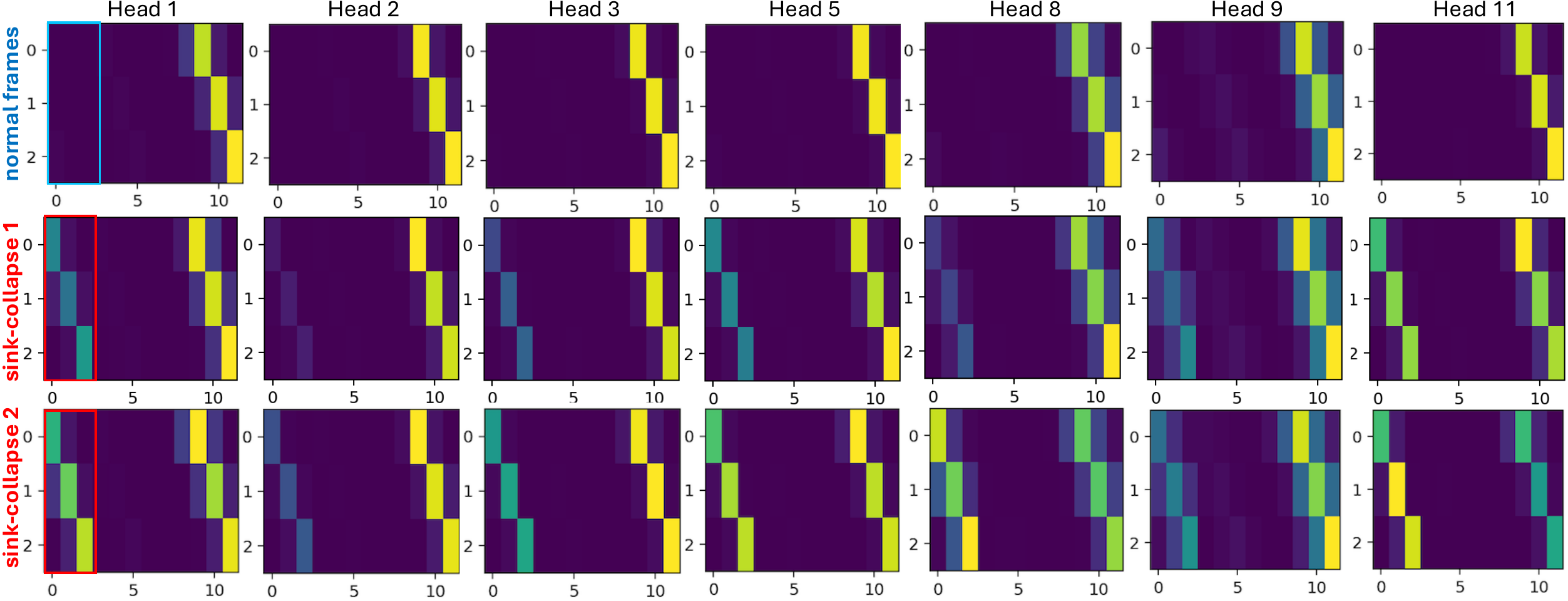}
    \caption{Visualization of inter-head attention homogenization. 
The first three latent frames serve as attention sinks, and the last three are being generated. 
Results are shown for the same DiT layer and diffusion step (KV cache size = 12, with 3 sink tokens) of different frames. 
The top row shows normal frames while the bottom two rows show two different sink-collapse frames where multiple attention heads simultaneously assign significantly higher weights to sink frames causing abrupt scene change back to these frames.}
    \label{fig:attention_heatmap}
\end{figure*}
In StreamingLLM, the initial tokens remain in the KV cache without being evicted. Similarly in autoregressive video generation, the initial frames get preserved when rolling the KV cache  which are referred to as sink frames. 

However, as shown in~\cref{fig:sink-collapse} that while sink frames can increase the overall alignment and stability, the generated videos are shown to constantly fallback to the initial frames abruptly, a phenomenon we refer to as sink collapse. For example, we plot the normalized L2 distance to the initial frames in~\cref{fig:sink-collapse} where significant drops at exactly the same positions such as frame index 132 and 201 can be seen across different prompts.

Upon close inspection, the key reason for sink collapse lies in the periodic nature of RoPE position embedding as describe before. While this rotation preserves relative phase differences in short contexts, the periodic trigonometric nature causes phase re-alignment at long horizons, effectively resetting positional distinctions. 
As generation proceeds autoregressively, this periodic aliasing leads multiple distant frames to share nearly identical embeddings, making the attention mechanism overemphasize these ``sink'' positions. Consequently, the model collapses into repetitive frames.

\subsection{Mitigating Sink Collapse}
\label{sec.mitigate}
The phenomenon of duplication has also been observed when extending bi-directional models to generate longer sequences as discussed in~\cite{zhao2025riflex} which proposes to identify the closest temporal dimension causing the repetition and alert its frequency. However, we find that it's ineffective in autoregressive generation. We show that there are two major distinctions between the repetition of bi-directional models and autoregressive models which are summarized below.

\textbf{Intra-head phase concentration} We find that the repetition is not caused by a single temporal dimension but the results of all dimension together which is shown  in~\cref{fig:sink-collapse} and~\cref{sec.ablation.single_dimension}. E.g, for the observed repeat frame index 132, RIFLEx predicts the closest intrinsic frequency component index to be 8 and the period of it to be 118 which deviates from the actual repetition by a large margin. In~\cref{sec.ablation.single_dimension}, we demonstrate that adjusting the closest intrinsic component alone does not resolve the issue, and that perturbing any individual RoPE component likewise yields no improvement, further justifying that the repetition is not caused by a single position embedding dimension. Instead, we visualize the phase concentration intensity in~\cref{fig:sink-collapse}. It can be observed that sink-collapse emerges almost exactly when the phase concentrations reach their local maxima. Formally, given the RoPE frequencies 
\(\omega_i = \theta_0^{-2i/d}\) for \(i = 1, \dots, K\), where \(K = \tfrac{d}{2}\),
we define the phase coherence kernel for a relative displacement 
\(\Delta = m - n\) as
\begin{equation}
C(\Delta)
= \Bigg|
    \frac{1}{K}
    \sum_{i=1}^{K}
    e^{\,j\,\omega_i \Delta}
  \Bigg|.
\label{eq:phase_kernel}
\end{equation}

Then the sink-relative phase concentration of a generated frame \(g\) can be formulated as \( R_{\text{sink}}(g) = C(g - s) \) where $s$ is the sink frame. A large value indicates that multiple RoPE frequency components become phase-aligned with the reference sink frame, leading to phase synchronization and consequently to the sink-collapse phenomenon.

\textbf{Inter-head attention homogenization} Prior work~\cite{shin2025motionstreamrealtimevideogeneration} has demonstrated that attending to sink frames constitutes a natural behavior that can enhance temporal alignment and stabilize the generation process. Nevertheless, given that modern transformer architectures rely on multi-head attention to capture diverse representational subspaces, our analysis reveals that sink-collapse is not attributed to any single attention head. Rather, it arises from a collective synchronization phenomenon, wherein multiple heads concurrently exhibit high phase concentration, resulting in a global degeneracy of attention diversity. We visualize the attention heatmap in~\cref{fig:attention_heatmap} for regular frames (top row) and  sink-collapse frames (bottom two rows) that's being generated in a chunk size of 3 which is used by both LongLive~\cite{yang2025longlive} and Self-Forcing++~\cite{cui2025self}. For regular frames, the model usually assign large self-attention weights to the tokens themselves that's being generated (last 3 frames) and distribute the weights evenly to the rest of the tokens. However, for the frames where sink-collapse happen, nearly all attention heads in the same layer assign significant large weights to both the frames that are being generated and the sink frames, causing the model to copy these frames from all sub-spaces created by multi-head attention and results in abrupt scene transition back to sink frames.

Based on these two observation, we propose a simple yet effective approach to significantly mitigate it. Since sink-collapse happens when all attention heads exhibit strong similarities to the sink frames simultaneously as shown in~\cref{fig:attention_heatmap}, we propose to shift the base frequency of different attention heads by a certain margin which we term as multi-head jitter which are described in the algorithm below. Owing to the inherent periodicity of the RoPE embedding, the introduced phase shift disrupts the global alignment among heads, thereby reducing the likelihood of simultaneous phase overlap across all heads and effectively mitigating the sink-collapse phenomenon.

\begin{algorithm}[H]
\caption{LoL: Multi-Head RoPE Jitter}
\label{alg:per_head_jitter_impl}
\KwIn{$Q,K\!\in\!\mathbb{R}^{B\times T\times H\times D}$, base $\theta_0$, jitter scale $\sigma_\theta$}
\KwOut{Rotated $Q',K'$ with head-wise perturbed frequencies}

\BlankLine
$\nu_i \!\leftarrow\! -2i/d_{\text{time}},\;\; i=0\!:\!D/2\!-\!1$ \tcp*{frequency exponents}

\For{$h = 1 \ldots H$}{
  $\epsilon_h \!\sim\! \mathcal{U}[-1,1]$ \hspace{2mm}$\hat\theta_h \!\leftarrow\! \theta_0(1+\boldsymbol{\sigma_\theta\epsilon_h})$\;
  
  $\omega_h \!\leftarrow\! [\,\hat\theta_h^{\nu_0},\ldots,\hat\theta_h^{\nu_{D/2-1}}\,]$\;
  
  $(Q'_h,K'_h) \!\leftarrow\! \mathrm{RoPERotate}(Q_h,K_h,\omega_h)$\;
}

$Q' \!\leftarrow\! [Q'_1,\ldots,Q'_H],\;\; K' \!\leftarrow\! [K'_1,\ldots,K'_H]$\;
\end{algorithm}

\subsection{Infinite  Streaming Generation}
\label{sec.infinite}
Besides sink-collapse, ultra-long video generation is also constrained by the length of RoPE and the memory consumption of VAE decoding. For instance, the maximum generation length of both LongLive~\cite{yang2025longlive} and Self-Forcing++~\cite{cui2025self} is 4 minutes and 15 seconds, limited by the 1024-frame latent sequence and VAE memory usage. We show that generation can be extended infinitely with little quality degradation owing to two inherent properties of current architectures: 1) Causal VAE. Both LongLive and Self-Forcing++ build upon Wan-2.1~\cite{wan2025}, which employs a 3D causal VAE that ensures temporal causality and allows a sliding-window decoding strategy which greatly reduces memory and computation. 2)  Local Attention. Both models adopt local attention over the most recent \(N\) latent frames to limit computational complexity. As shown in~\cref{eq:rope_relative}, the dot product of two RoPE~\cite{su2024roformer} embeddings mainly depends on their relative positional difference.  

Thus, with sink-collapse mitigated, the model can generate videos of infinite length. During streaming generation, both initial noise and RoPE will be dynamically sampled, introducing minimal additional overhead compared to pre-generated methods.

\section{Experiments}
\begin{figure*}
    \centering
\includegraphics[width=1.0\linewidth]{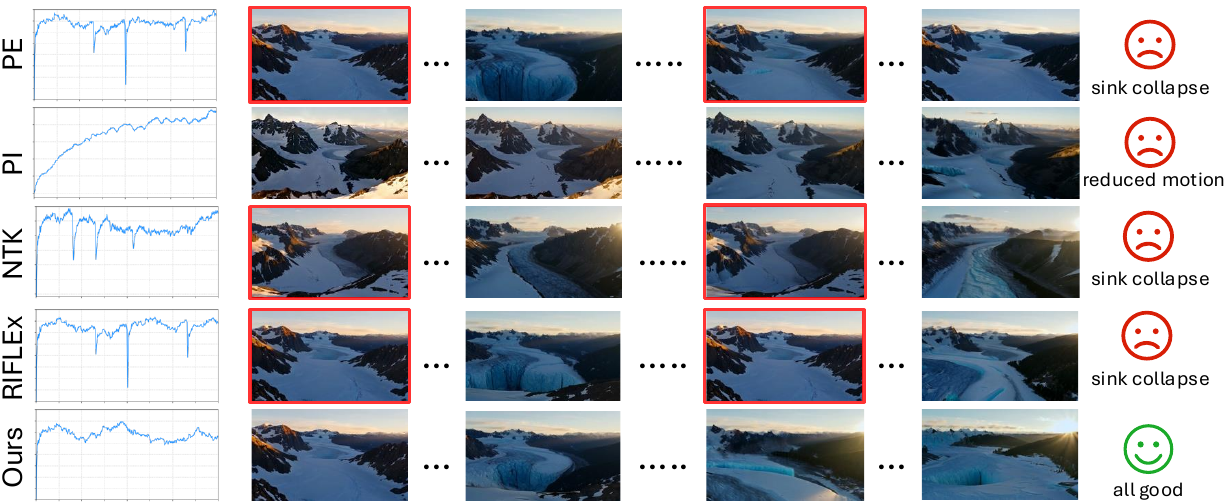}
    \caption{Visualization of the results after applying different PE extension methods. Baseline approaches continue to exhibit sink-collapse or diminished motion, whereas our method effectively alleviates sink-collapse and best preserves motion dynamics, as shown in~\cref{tab:pe_performance_overall}.}
    \label{fig:main_figure}
\end{figure*}
\label{sec:experiments}

\subsection{Settings}
\label{sec.settings}
\textbf{Baseline methods} Following RIFLEx~\cite{zhao2025riflex}, which tackles repetition in bidirectional models, we compare with several methods addressing position embedding limitations. PE directly extends the sequence length without modifying embeddings and serves as the baseline. PI~\cite{chen2023extending} rescales RoPE frequencies as $\theta_{\mathrm{PI}} = \theta / s$, where $s = L'/L$ and $L$ and $L'$ denote the training and inference sequence lengths, respectively. NTK~\cite{bloc97} adjusts the base frequency $b$ such that each dimension's frequency is scaled as $(\lambda b)^{-2j/d'}$, where $\lambda = s d' / (d' - 2)$, $d'$ is the rotary dimension, and $s = L'/L$ is the same as above. YARN~\cite{peng2023yarn} introduces fine-grained frequency grouping based on the number of cycles $r_j = (2\pi)^{-1} L \theta_j$ over the training length $L$ and interpolates between original and scaled frequencies using thresholds $\alpha$ and $\beta$ via $\theta^{\text{YaRN}}_j = \gamma(r_j)\theta_j + (1-\gamma(r_j))\theta_j/s$. Finally, RIFLEx~\cite{zhao2025riflex} identifies the intrinsic repetition frequency
$k = \arg\min_j |N_j - N|$ with $N_j = 2\pi / \theta_j$, and updates it to $\theta_k = 2\pi / (L s)$ to ensure $N'_k \geq L s$, effectively mitigating repetition in bidirectional models such as HunyuanVideo~\cite{li2025hunyuan} and CogVideoX~\cite{hong2022cogvideo}.

Additionally, we compare with the following autoregressive methods on general video generation benchmarks such as NOVA~\citep{deng2024nova}, Pyramid Flow~\citep{jin2024pyramidal}, SkyReels-V2-1.3B~\citep{chen2025skyreels}, MAGI-1-4.5B~\citep{teng2025magi}, CausVid~\citep{yin2025causvid} and Self-Forcing~\citep{huang2025self}.

\textbf{Evaluation metrics}
In order to measure sink-collapse, we adopt No-Repeat score from RIFLEx~\cite{zhao2025riflex} which computes the normalized L2 distance to the initial frames and was used to measure the repetition problem in bidirectional models. Here, we apply a more strict measurement and compute the scores in the whole generation range and take the maximum value. Specifically, we compute two scores including Sink-Collapse Max which computes the maximum distance drop with respect to sink frames in any single test prompt,  indicating the worst case scenario and Sink-Collapse Avg which computes the average distance drop across all test prompts, indicating the average scenario. Also following~\cite{li2025stable,yin2025causvid,huang2025self,cui2025self}, we  adopt VBench ~\citep{huang2024vbench} to evaluate the quality of generated videos~\cite{yin2025causvid,cui2025self}, consisting of 128 prompts from MovieGen~\citep{polyak2024movie}. 
We expect the proposed method to significantly improve Sink-Collapse score while maintaining similar generation quality.

\textbf{Implementation Details}
We test our method on state-of-the-art ultra long video generation models including LongLive~\cite{yang2025longlive} and Self-Forcing++~\cite{cui2025self} with a local attention window size of 12 and sink frames of 3 which uses 1.3B base model that enables generation at 20 fps on a single NVIDIA H100 as shown in~\cite{yang2025longlive}. For baseline position embedding extension methods, we use the last frame index before the first sink-collapse, e.g, 132 as the original length denoted as $L$ in our equations and the expected generation length as the output length denoted as $L'$ in our equations. The frame indices in this paper all refer to latent frames, e.g, latent index 132 corresponds to around 528 video frames which is around 33 seconds at a frame rate of 16.

\begin{table}
\centering
\caption{Results of applying our methods to LongLive and Self-Forcing++ with attention sink frames on videos of 100 seconds. In both cases, our method can effectively mitigate the sink collapse problem while maintaining the overall quality of the video.}

\begin{threeparttable}
\resizebox{\textwidth}{!}{
\begin{tabular}{lccc|ccccccc}
\toprule
\multirow{3}{*}{Method}&\multicolumn{3}{c}{Core Metrics} & \multicolumn{7}{c}{General Metrics}\\
\cmidrule(lr){2-4}\cmidrule(lr){5-11}
 & Sink-Collapse & Sink-Collapse & Dynamic & Temporal & Text & Clip & Subject & Background & Motion & Imaging \\
                & Max $\downarrow$    & Avg$\downarrow$     & Degree$\uparrow$  & Quality   & Alignment & Score & Consistency & Consistency & Smoothness & Quality \\
\hline
\rowcolor{skyblue!30}\multicolumn{11}{c}{\textit{Results on LongLive}}\\
PE & \cellcolor{red!10}73.06 & \cellcolor{red!10}30.54 & \cellcolor{green!10}34.62 & 88.56 & 28.09 & 31.91 & 97.53 & 96.21 & 98.91 & 69.59 \\
PI & \cellcolor{green!10}4.97 & \cellcolor{green!10}2.27 & \cellcolor{red!10}00.35 & 85.25 & 24.62 & 29.11 & 99.65 & 99.00 & 99.00 & 56.47 \\
NTK & \cellcolor{red!10}41.11 & \cellcolor{red!10}11.64 & \cellcolor{green!10}28.72 & 87.95 & 27.98 & 31.79 & 97.88 & 96.57 & 99.03 & 69.83 \\
YARN & \cellcolor{green!10}11.17 & \cellcolor{green!10}5.08 & \cellcolor{red!10}2.67 & 85.17 & 28.02 & 31.93 & 99.35 & 97.93 & 99.49 & 68.89 \\
RIFLEX & \cellcolor{red!10}70.95 & \cellcolor{red!10}29.93 & \cellcolor{green!10}35.11 & 88.61 & 23.11 & 28.38 & 97.48 & 96.22 & 98.89 & 69.47 \\
Ours (LoL) & \cellcolor{green!10}16.67 & \cellcolor{green!10}3.93 & \cellcolor{green!10}35.27 &88.69 & 27.80 & 31.68 & 97.62 & 96.25 & 98.91 & 69.45 \\
\hline
\rowcolor{skyblue!30}\multicolumn{11}{c}{\textit{Results on Self-Forcing++}}\\
PE & \cellcolor{red!10}68.07 &\cellcolor{red!10} 34.11 & \cellcolor{green!10}83.32 & 93.14 & 27.62 & 31.36 & 93.41 & 93.13 & 97.79 & 63.06 \\
PI & \cellcolor{green!10}17.07 & \cellcolor{green!10}2.62 & \cellcolor{red!10}1.95 & 84.66 & 26.51 & 31.05 & 98.94 & 97.19 & 99.19 & 69.80 \\
NTK & \cellcolor{red!10}49.65 & \cellcolor{red!10}14.96 & \cellcolor{green!10}82.90 & 93.04 & 27.26 & 30.95 & 93.41 & 93.10 & 97.68 & 62.86 \\
YARN & \cellcolor{green!10}33.04 & \cellcolor{green!10}6.69 & \cellcolor{red!10}36.71 & 88.29 & 27.19 & 31.26 & 96.77 & 95.28 & 95.28 & 67.50 \\
RIFLEX & \cellcolor{red!10}66.56 & \cellcolor{red!10}32.86 & \cellcolor{green!10}82.36 & 93.01 & 27.57 & 31.32 & 93.48 & 93.10 & 97.79 & 63.26 \\
Ours (LoL) & \cellcolor{green!10}22.70 & \cellcolor{green!10}6.12 & \cellcolor{green!10}81.20 & 92.91 & 27.38 & 31.14 & 93.74 & 93.11 & 97.72 & 62.92 \\
\bottomrule
\end{tabular}
}
\begin{tablenotes}
\footnotesize
\item \colorbox{red!10}{Red color} and \colorbox{green!10}{Green color} indicates whether there is severe repetition or significantly reduced motion. 
\item $\downarrow$ indicates lower is better, $\uparrow$ indicates higher is better. 
\end{tablenotes}
\end{threeparttable}

\label{tab:pe_performance_overall}
\end{table}
\begin{table}[t]
  \centering
  \caption{Performance comparisons with other autoregressive video generation models on 75s and 100s videos. Results show that despite being training-free, our method effectively addresses the sink-collapse problem while maximally preserves video generation quality.}  
  \resizebox{1.0\textwidth}{!}{
\begin{tabular}{lcccc|cccc}
  \toprule
  \multirow{3}{*}{Model} &
  \multicolumn{4}{c}{Results on 75s $\uparrow$} &
  \multicolumn{4}{c}{Results on 100s $\uparrow$} \\
  \cmidrule(lr){2-5} \cmidrule(lr){6-9}
   & Text & Temporal & Dynamic &  Framewise
   & Text & Temporal & Dynamic  & Framewise \\
   & Alignment & Quality & Degree  & Quality
   & Alignment & Quality & Degree& Quality \\
  \midrule
  \rowcolor{skyblue!30}
  \multicolumn{9}{l}{\textit{Autoregressive models}}\\
  NOVA        & 23.37 & 86.32 & 31.24  & 31.53  & 22.89 & 86.24 & 31.09 & 31.03 \\
  MAGI-1    & 24.95 & 87.89 & 24.82&  52.04  & 23.75 & 87.62  & 22.21 & 50.90 \\
  SkyReels-V2  &  22.70 & 88.99 & 39.89  & 51.55  & 22.05 & 88.80 & 38.75  & 50.48 \\
  CausVid      & 24.76 & 89.14  & 35.82 & 60.96  & 24.41 & 89.06 & 34.60& 61.01 \\
  Self Forcing & 23.39 & 87.79 & 29.15 & 60.02  & 22.00 & 87.39 & 26.41& 58.25 \\
  \rowcolor{skyblue!30}
  \multicolumn{9}{l}{\textit{Ultra-Long Autoregressive models}}\\
  Self-Forcing++  &
                 26.31 & 91.00  & 55.62 & 60.67 &
                 26.04 & 
                  90.87&54.12  & 60.66 \\
  LongLive  &
                 \textbf{28.08} & 88.64  & 35.14  & \textbf{64.16} &
                 \textbf{28.09} & 
                  88.56&34.62  & \textbf{64.05} \\
  Self-Forcing++(LoL)&27.39&\textbf{93.08}&\textbf{81.30}&60.59&27.38&\textbf{92.91}&\textbf{81.20}&60.25\\
  LongLive(LoL)  &
                 27.85 & 88.78  & 35.77 & 63.90 &
                 27.80 & 
                  88.69 &35.27  & 63.76 \\
  \bottomrule
\end{tabular}
\vspace{-0.2cm}
}

  \label{tab:auto_baseline}
\end{table}
\subsection{Empirical Results}
We present quantitative results of applying our method to LongLive~\cite{yang2025longlive} and Self-Forcing++~\cite{cui2025self} in~\cref{tab:pe_performance_overall,tab:auto_baseline}. As shown, naive position extrapolation (PE) leads to severe sink-collapse in both models. For instance, in LongLive, the maximum distance drop reaches 73.06 and the average drop 30.54; in Self-Forcing++, these values reach 68.07 and 34.11, respectively. This demonstrates that despite their distinct training paradigms, both methods suffer from pronounced sink-collapse when using attention sink frames. By contrast, position interpolation (PI) effectively alleviates the issue since positional encodings are interpolated before the first repetition occurs. However, this comes at the cost of a drastic reduction in dynamic degree where video motion becomes almost stagnant, consistent with observations in~\cite{zhao2025riflex}.  More advanced approaches such as NTK and YARN each make different trade-offs. NTK maintains higher motion dynamics with only a minor drop (about 6\% for LongLive) but offers limited mitigation against sink-collapse. YARN, on the other hand, strongly suppresses sink-collapse but greatly hampers dynamics, similar to position interpolation. RIFLEx achieves state-of-the-art performance in reducing repetition for bidirectional models while largely preserving motion under the autoregressive setting. However, since it attributes repetition to a single temporal dimension, it fails to address sink-collapse effectively with collapse scores similar to position extrapolation. In contrast, our method (LoL) substantially mitigates sink-collapse in both maximum and average scores, reaching levels comparable to position interpolation, while simultaneously preserving motion dynamics similar to position extrapolation. As a result, LoL enables indefinite, streaming video generation without collapsing, as illustrated in~\cref{fig:main_figure}.

Additionally, we provide comparison results with other autoregressive baseline methods in~\cref{tab:auto_baseline} which shows that our  method can effectively address the sink-collapse problem without hindering the generation quality.

\subsection{Ablation Study}
\subsubsection{Is repetition dominated by a single dimension?}
\label{sec.ablation.single_dimension}
\begin{figure}[H]
    \centering
    \begin{subfigure}[t]{0.48\linewidth}
        \centering
        \includegraphics[width=\linewidth]{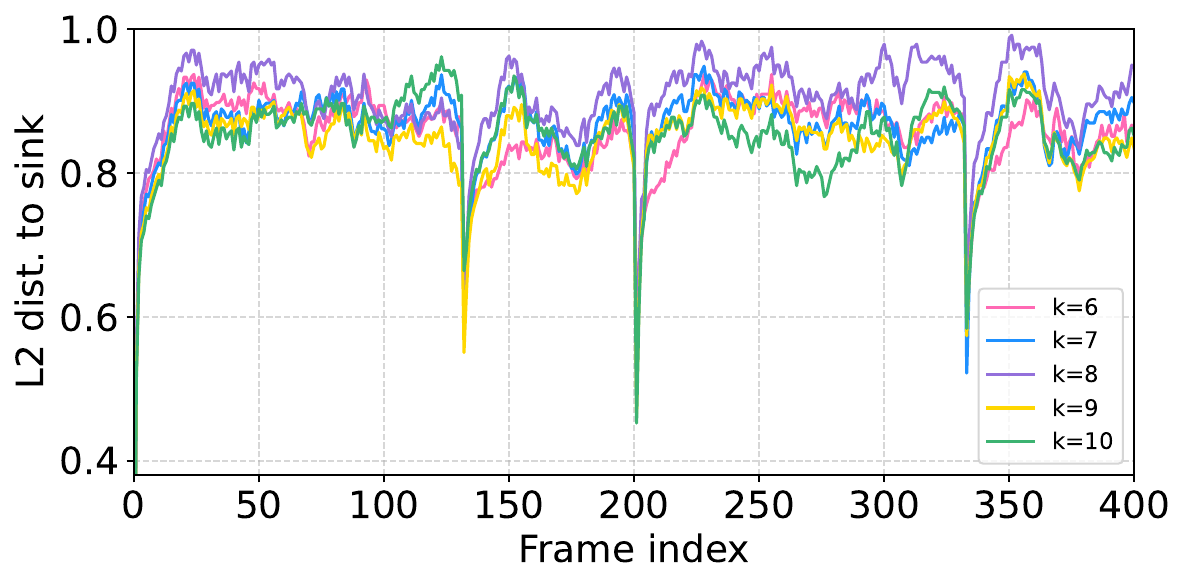}
        \caption{Normalized \(L_2\) distance to sink frames of modifying the frequencies of the dimension identified by RIFLEx and its neighboring dimensions. None of the modifications alone can mitigate the sink-collapse phenomenon.}
        \label{fig:riflex}
    \end{subfigure}\hfill
    \begin{subfigure}[t]{0.48\linewidth}
        \centering
        \includegraphics[width=\linewidth]{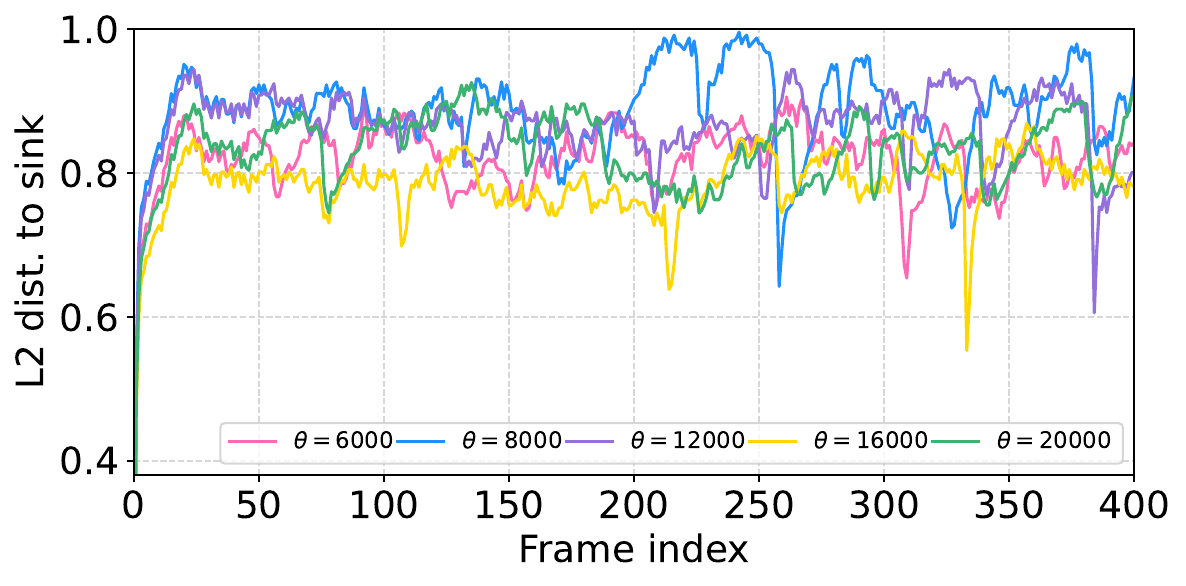}
        \caption{Normalized \(L_2\) distance to the first frame under different RoPE base values \(\theta\). Varying \(\theta\) shifts the collapse index but fails to address the underlying problem.}
        \label{fig:ablation_base}
    \end{subfigure}
    \caption{Ablation studies on sink-collapse behavior under frequency and RoPE base modifications.}
    \label{fig:ablation_combined}
\end{figure}

Repetition phenomenon has been studied in bi-directional models as well. As mentioned in ~\cref{sec.settings} that state-of-the-art work such as RIFLEx~\cite{zhao2025riflex} tries to identify the closest dimension that approximates the  observed duplication frame and solve the problem by changing the frequency of that dimension. However, as we have shown before that sink-collapse in autoregressive generation is not caused by a single dimension but the results of all dimensions together in~\cref{fig:sink-collapse}. Here, we further show that not only does adjusting the closest intrinsic dimension fail, but altering the frequency of any position-embedding dimension is also ineffective. We plot the results of altering the dimension identified by RIFLEx and its neighboring dimensions in~\cref{fig:riflex}.

It can be seen that regardless of the dimensions whose frequency are altered, the sink-collapse cannot be effectively mitigated which further proves that the repetition is not caused by a single dimension.

\subsubsection{Can sink-collapse get fixed by different RoPE base?}
In our main experiments, we follow the base mode~\cite{wan2025wan,yang2025longlive,cui2025self} and use the standard RoPE base value of 10{,}000. While one might consider adjusting the base~$\theta$ to alleviate the sink collapse issue, our analysis shows that this modification alone does not fundamentally resolve the problem. Changing~$\theta$ only alters the phase progression rate of the positional embeddings, shifting the point at which sink collapse occurs rather than eliminating it. In summary, varying~$\theta$ merely delays or advances the collapse schedule along the temporal axis. As shown in~\cref{fig:ablation_base}, we set the base from 6000 to 20,000 which are significantly different from the original base value. However, the sink collapse phenomenon is not mitigated but shifted for different bases. In contrast, our method directly mitigates the root cause of sink collapse as described in~\cref{sec.mitigate}, maintaining stable and consistent generation quality across extended temporal ranges.

\subsubsection{What's the impact of different jitter intensity?}
\begin{figure}[t]
    \centering
    \begin{subfigure}[t]{0.48\linewidth}
        \centering
        \includegraphics[width=\linewidth]{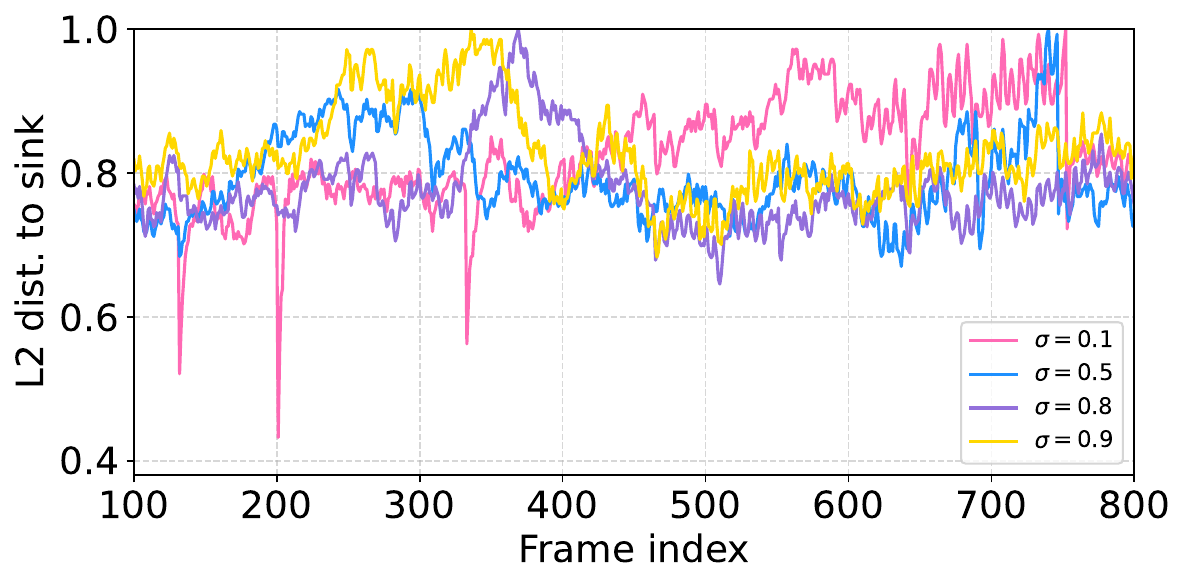}
        \caption{Normalized \(L_2\) distance to sink frames for different RoPE jitter \(\sigma\). Empirically, a value of 0.8 achieves a good balance between generation quality and sink-collapse mitigation.}
        \label{fig:ablation_jitter}
    \end{subfigure}\hfill
    \begin{subfigure}[t]{0.48\linewidth}
        \centering
        \includegraphics[width=\linewidth]{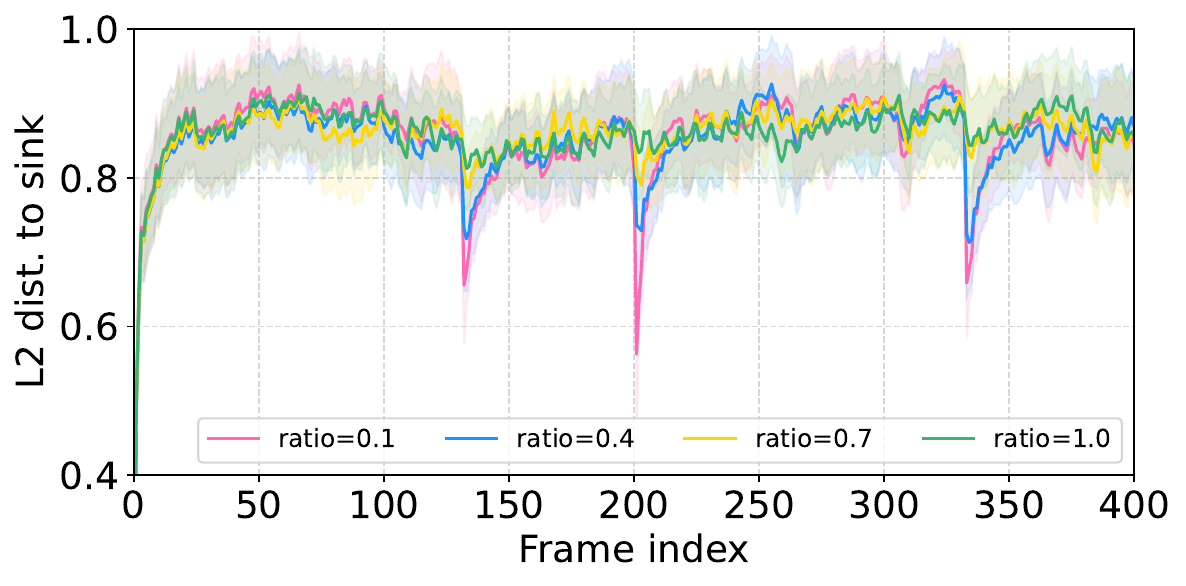}
        \caption{Normalized \(L_2\) distance to sink frames with varying numbers of jitter attention heads, averaged over three random seeds. Increasing the number of jitter heads gradually alleviates the sink-collapse phenomenon.}
        \label{fig:ablation_jitter_head_num}
    \end{subfigure}
    \caption{Ablation studies on RoPE jitter magnitude and the number of jitter attention heads.}
    \label{fig:ablation_jitter_combined}
\end{figure}

Here we show the results of applying different jitter intensity $\sigma$. As shown in~\cref{fig:ablation_jitter} that as we increase $\sigma$, the sink collapse phenomenon starts to alleviate. E.g, when the jitter intensity is 0.1 which is very small, the model still suffers from significant sink-collapse problem. 

When we set $\sigma$ to 0.5, the severity of sink-collapse has reduced, however, the problem can still be seen in extended generation length which can be seen at round 750 frame index. As we further increase the value to 0.8, the figure shows a smooth transition between frames and no significant drops are seen. Note that further increasing $\sigma$ can also mitigate the problem but at the cost of reduced motion or quality. In general, we find 0.8 to strike a good balance between generation quality and mitigating sink-collapse which can be seen from the quantitative results in~\cref{tab:pe_performance_overall} and ~\cref{tab:auto_baseline}.
\subsubsection{How many attention heads do we need to jitter?}
We randomly sample the number of heads to jitter based on the specified ratio using three different random seeds and report the aggregated results with standard deviation in~\cref{fig:ablation_jitter_head_num}. As illustrated, increasing the number of jittered heads progressively alleviates the sink-collapse phenomenon, further confirming that the issue does not originate from any single attention head. Empirically, jittering all heads achieves the most significant mitigation while maintaining high generation quality, as shown in~\cref{tab:pe_performance_overall,tab:auto_baseline}.

\section{Limitations and Further work}
Although our method is training-free, fine-tuning or retraining may further improve overall performance. The generation quality is constrained by the underlying models, which rely on local attention and sink frames for improved alignment and stability. However maintaining long-term memory still remains challenging, especially for multi-hour videos which is an important direction of our future work.
Given the inherent periodicity of RoPE position embeddings, future research may explore alternative embedding schemes or  advanced training strategies.
We also plan to improve controllability by integrating stronger control signals~\cite{wang2025ati,shin2025motionstreamrealtimevideogeneration}, and enhance scalability by incorporating sparse or linear attention mechanisms~\cite{chen2025sana,team2025longcat}.
\section{Limitations and Further work}
As shown previously, the repetition doesn't necessarily happen at global maximum of the RoPE phases. It happens at local maximums. Thus, our method can only significantly suppress the repetition behavior where sink collapse are rarely seen during hour-long video generation. Future work should investigate non-periodic position embeddings or advanced training strategies such as masking the sink frames should a repetition occurs. Although our method shows the potential for infinite video generation with little quality degradation, it is based on Wan2.1-T2V-1.3B, which has limited capacity. In addition, since the model has never been trained on real datasets, its performance is bounded by that of the teacher model. We plan to explore the use of stronger base models in the future. Currently, the model can generate videos in a streaming manner at about 16 FPS on a single GPU. We also hope to further improve its efficiency and extend it to support control signals in future work.

\clearpage

\bibliographystyle{plainnat}
\setlength{\bibhang}{0pt}   
\setlength\bibindent{0pt}   
\bibliography{main}

\clearpage


\vspace{1.5em}

\makeatletter
\if@twocolumn
  \onecolumn
\fi
\makeatother

\section*{More related work}
\textbf{Video Diffusion Models}
The landscape of video generation has been profoundly reshaped by the emergence of the Diffusion Transformer (DiT)~\cite{peebles2023scalable}, whose performance scales remarkably with computational capacity. Building upon DiT, a line of breakthroughs has followed. Sora~\citep{openai2024sora} delivers highly realistic and temporally consistent videos with diverse, natural motion. Wan 2.1~\citep{wan2025} highlights the power of large-scale pretraining for high-resolution video synthesis, while CogVideoX~\citep{hong2022cogvideo,yang2024cogvideox} enhances cross-modal fusion through expert transformers and adaptive LayerNorm, achieving superior text–motion alignment. Different from these architectures, Hunyuan Video~\citep{kong2024hunyuanvideo} introduces a causal 3D VAE~\citep{kingma2013auto} for efficient spatio-temporal token compression and couples it with a large language model for text conditioning, yielding outstanding results. Commercial systems such as Seedance~\cite{gao2025seedance} and Kling~\cite{klingaiKlingNextGen} push quality and efficiency to new heights, while open initiatives like Open-Sora~\citep{lin2024open,peng2025open} and Open-Sora-Plan~\citep{lin2024open} bring these advances to the open-source community, greatly accelerating progress in both realism and scalability.

\section*{Phase alignment of other $\theta$ values}
\begin{figure}[H]
    \centering
    \includegraphics[width=1.0\linewidth]{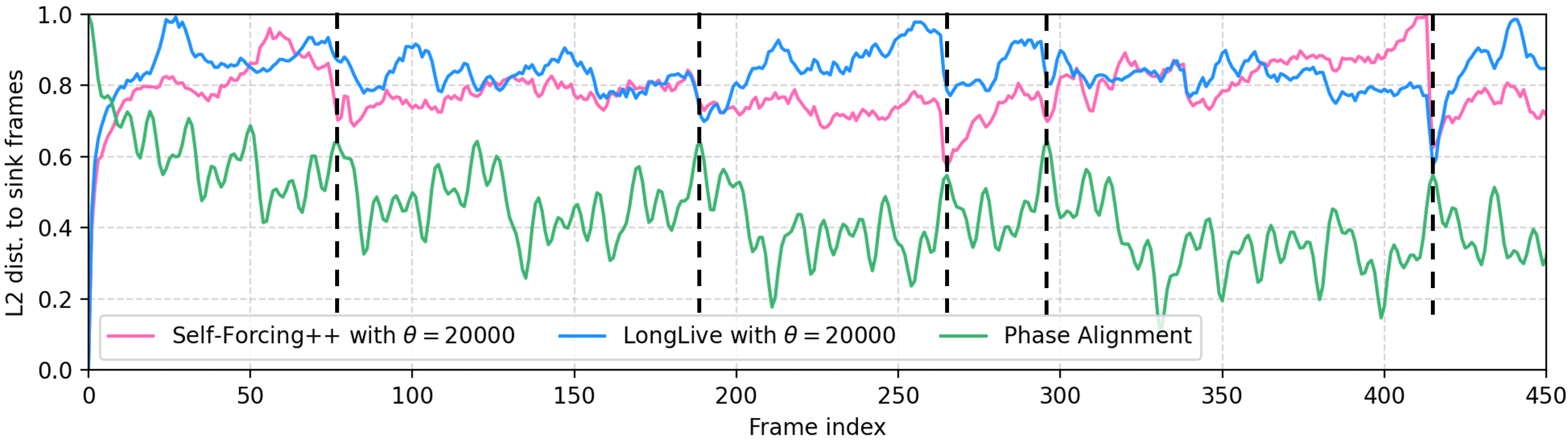}
    \caption{The L2 distance to sink-frames compared to phase alignment at $\theta$=20000. It can be seen that the sink-collapse location also highly correlates with the local maxima of phase concentration.}
    \label{sup.fig:different_base_prediction}
\end{figure}

In this paper, our primary focus is on analyses conducted with the RoPE base value $\theta = 10000$, which is the default configuration used in the pretrained Wan2.1-T2V-1.3B model~\cite{wan2025}. This setting is also adopted by both LongLive~\cite{yang2025longlive} and Self-Forcing++~\cite{cui2025self}, where the same base value is retained in order to remain consistent with the original positional encoding spectrum of the backbone model. Here we further examine the behavior of sink-collapse under alternative choices of the RoPE base and analyze their patterns with respect to phase concentration. As illustrated in~\cref{sup.fig:different_base_prediction}, experiments with a different $\theta$ value reveal a highly consistent pattern similar to that of $\theta=10000$. Sink-collapse phenomeon remains closely aligned with the emergence of phase concentration. In particular, the normalized L2 distance between consecutive representations shows a clear drop at the time step where the phase alignment reaches its local maxima. This provides additional empirical evidence for the problem discussed in~\cref{sec.mitigate}.

\section*{The impact of sink frame number}
In the main paper, our experiments primarily focus on the configuration that applies three sink tokens together with a local attention window of nine, resulting in a total window size of twelve. This setting follows the default configuration used in~\cite{yang2025longlive}. To further examine the robustness of the sink-collapse phenomenon, we additionally investigate how the number of sink frames affects the model's behavior.

\begin{figure}[H]
    \centering
    \includegraphics[width=\linewidth]{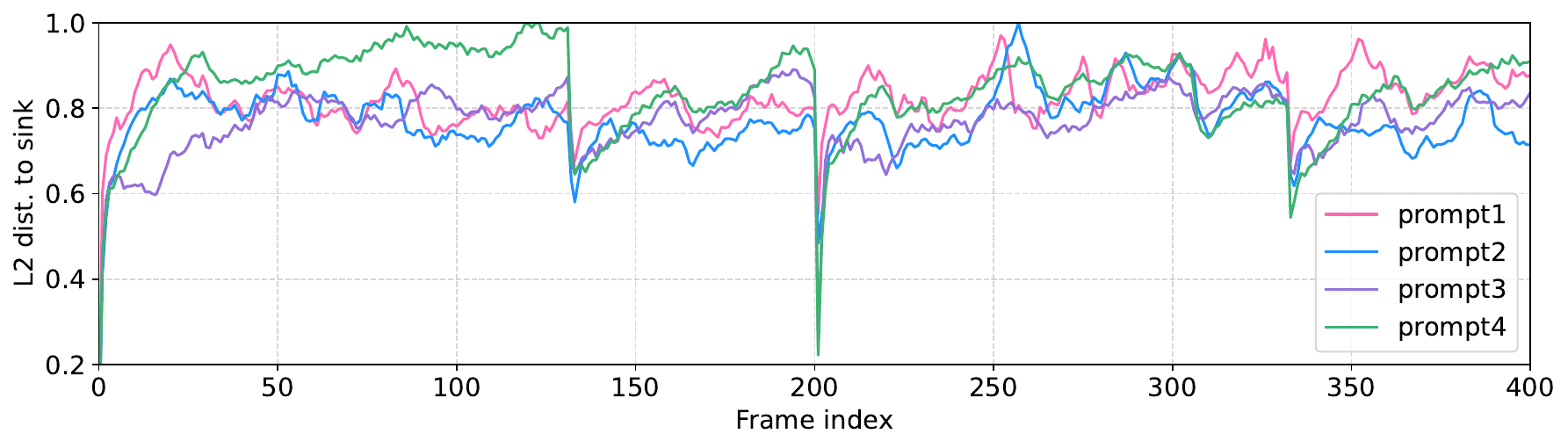}
    \caption{Behavior of the model when configured with five attention-sink frames. Despite allocating a larger sink region, the generation still fall into repetitive patterns, illustrating that increasing the sink-frame budget does not alleviate the collapse.}
    \label{fig:sink_frame_5}
\end{figure}

As shown in~\cref{fig:sink_frame_5}, we visualize the results obtained when increasing the number of sink frames from three to five. The results indicate that sink-collapse continues to occur even with a larger number of sink frames. This observation suggests that the instability is not sensitive to the specific choice of sink frame count and that increasing the number of sink frames alone does not resolve the underlying issue.

\begin{figure}[H]
    \centering
    \includegraphics[width=\linewidth]{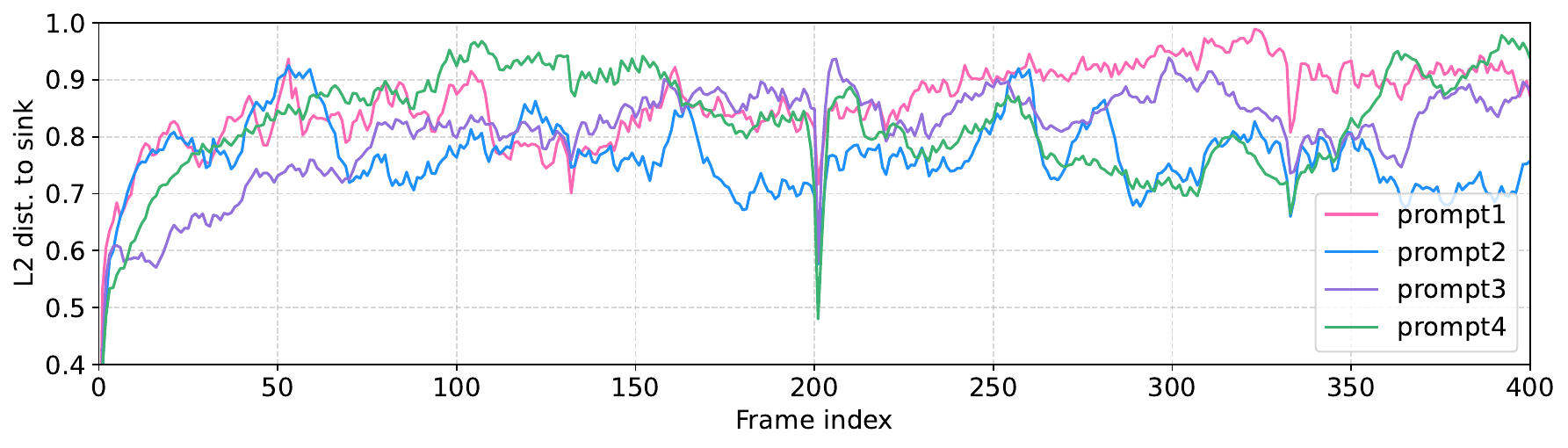}
    \caption{Visualization of the sink-collapse phenomenon when only one attention-sink frame is used. The results indicate that reducing the sink-frame size to its minimum does not eliminate repetition artifacts, and the model still exhibits signs of sink collapse.}
    \label{fig:sink_frame_1}
\end{figure}

Next, we reduce the number of sink frames to one, which represents the minimal possible sink-frame configuration. As shown in~\cref{fig:sink_frame_1}, even under this extreme setting, the model continues to exhibit sink-collapse artifacts, indicating that the phenomenon persists despite minimizing the number of designated sink frames.

\section*{Visualization of other model layers}
\begin{figure}[H]
    \centering
    \includegraphics[width=\linewidth]{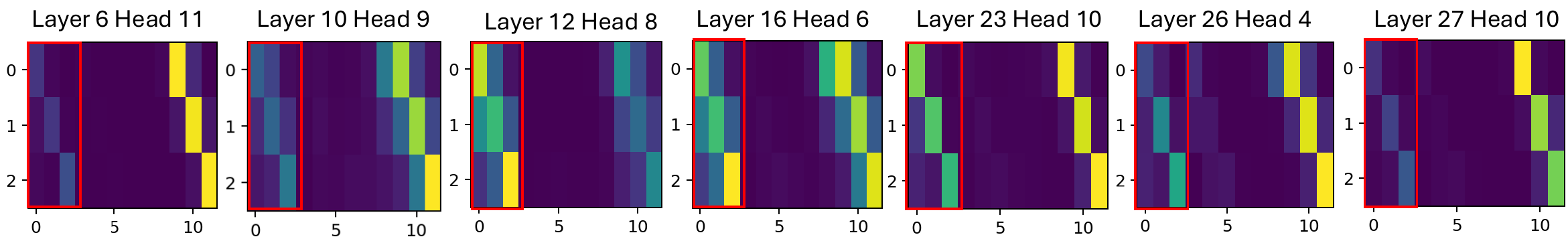}
    \caption{More visualization of attention heat map across different layers and different heads. It can be seen that the copying behavior is a prevalent issues throughout the model layers and attention heads.}
    \label{fig:more_attention_map}
\end{figure}

Beyond the visualization provided in~\cref{fig:attention_heatmap}, we further include additional attention maps extracted from several different attention layers at the identified sink-collapse locations, as shown in~\cref{fig:more_attention_map}. These  visualizations demonstrate that the copying behavior is not confined to a single layer or a specific head. Instead, it consistently appears across multiple attention layers and is exhibited by a wide range of attention heads. This widespread pattern reinforces the observation that sink-collapse is a structural effect emerging throughout the model rather than a localized anomaly. Consequently, as described in the algorithm, shifting the phase alignment serves as an effective strategy to mitigate this behavior by disrupting the conditions under which such copying behavior becomes dominant.

\section*{Visualization of single prompt }
\begin{figure}[H]
    \centering
    \includegraphics[width=\linewidth]{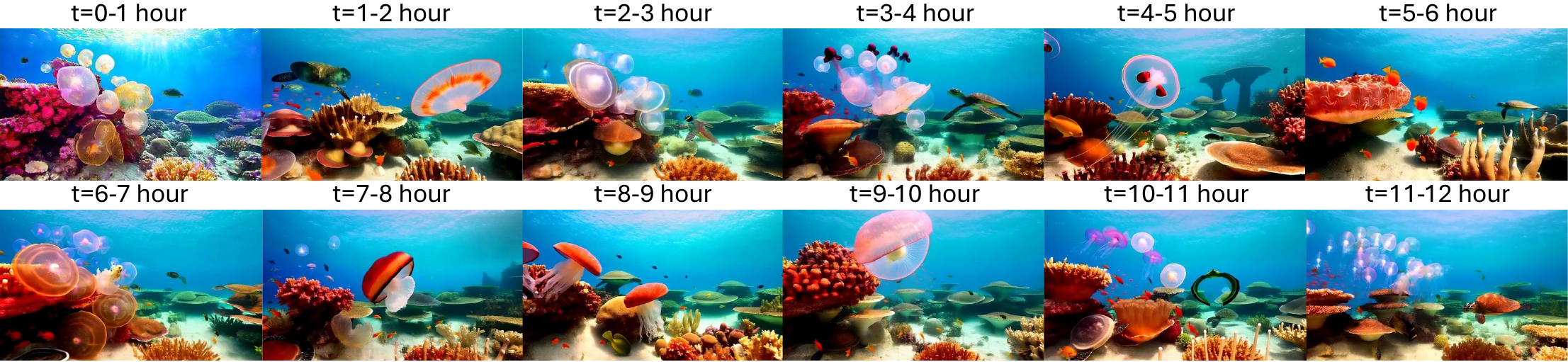}
    \caption{12-hour visualization for streamingly generated video for prompt "An enormous bloom of radiant jellyfish drifts through a breathtaking underwater world..."}
    \label{fig:jelly_fish_12_hours}
\end{figure}

\begin{figure}[H]
    \centering
    \includegraphics[width=\linewidth]{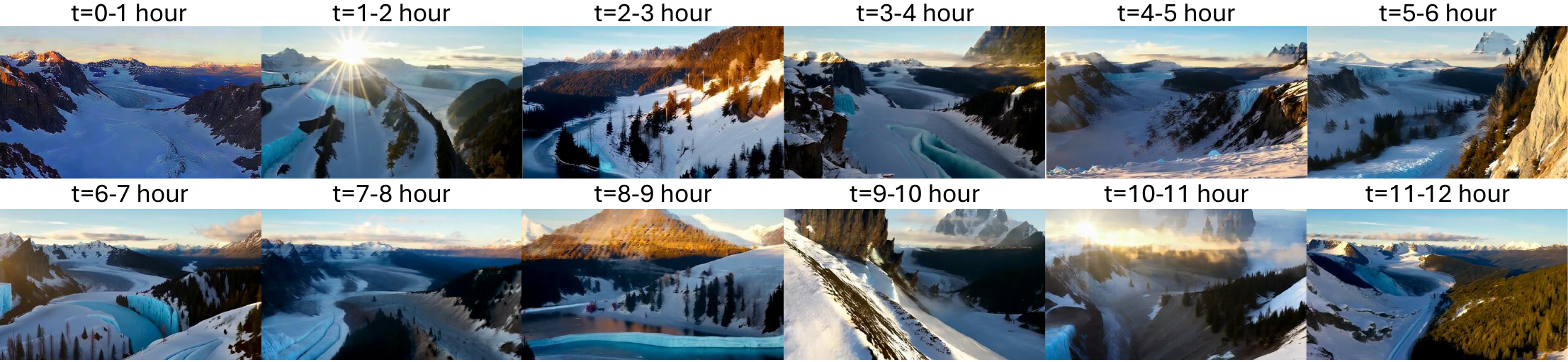}
    \caption{12-hour visualization for streamingly generated video for prompt "Cinematic FPV aerial shot flying forward over snow-capped mountains at golden hour, skimming along a razor ridgeline then dipping into a glacier valley..."}
    \label{fig:snow_mountain_12_hour}
\end{figure}

Here, we present additional long-form visualizations of our generated videos, extending up to 12 hours in duration, as illustrated in~\cref{fig:jelly_fish_12_hours}. These extended results further demonstrate the robustness of our approach that the model sustains coherent structure, visual consistency, and motion stability throughout the entire generation window, with little quality degradation. This further proves that that video generation method using local attention and  RoPE~\cite{su2024roformer} can generate to much longer sequence as described in~\cref{sec.infinite}.
\noindent Below is another streamingly generated 12-hour video featuring noticeably faster and more dynamic camera motion, as illustrated in~\cref{fig:snow_mountain_12_hour}. Despite the increased motion complexity, the generated sequence remains coherent and temporally consistent throughout the entire duration. This further demonstrates that our method maintains stable performance even under long-range, fast-motion scenarios.

\section*{Visualization of prompt switching}
\begin{figure}[H]
    \centering
    \includegraphics[width=\linewidth]{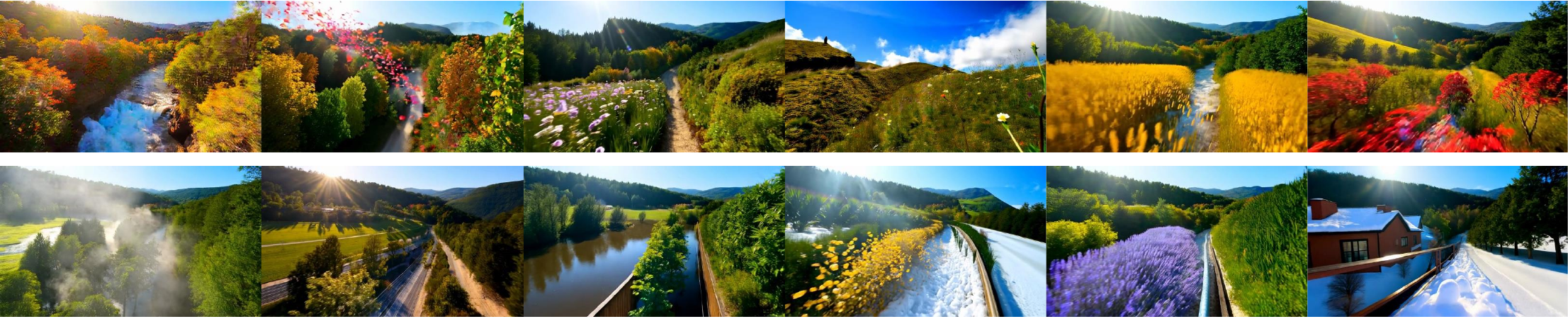}
    \caption{1-hour visualization for streamingly generated video for prompts ``The lens floats through a valley of multicolored trees'' followed by ``Leaves of emerald, amber, and rose swirl through the air like confetti.'', ``Motion forward through a valley of yellow-colored grass'', etc and 12 hours for single prompt ``A cinematic third-person shot of a wingsuit flyer racing through a  mountain valley...''.}
    \label{fig:multi-color-tree}
\end{figure}

\begin{figure}[H]
    \centering
    \includegraphics[width=\linewidth]{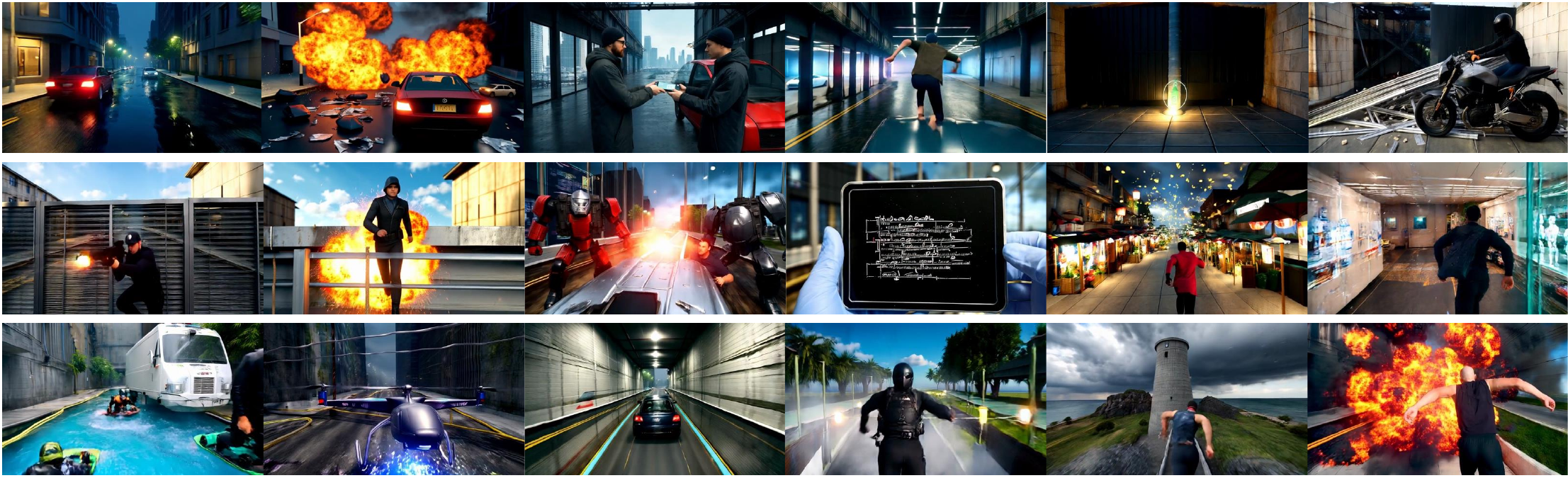}
    \caption{10-minute visualization with large scene transition for streamingly generated video for prompts ``Nighttime streets blaze with neon on rain-slick asphalt as the camera hurtles into a high-speed chase.'', ``The chase bursts onto the freeway. A fuel truck erupts, hurling flaming debris while the cars knife through shockwaves and falling metal.'', ``A massive explosion tears through the street as the hero leaps through flame and shrapnel toward the finale.'', etc}
    \label{fig:car_chase}
\end{figure}

Our method is compatible with both single-prompt generation and scenarios that require prompt switching. To demonstrate this capability, we first present a one-hour video generated with several controlled prompt transitions in~\cref{fig:multi-color-tree}, where the scene evolves as the prompts change over time. We further include another ultra-long video lasting 10-minutes but with a substantially larger number of prompt transitions in~\cref{fig:car_chase}, highlighting a much more challenging setting with frequent semantic shifts. Across both examples, the generated sequences remain coherent and stable, indicating that our approach scales effectively even as the number and frequency of prompt switches increase.
\noindent 

\section*{Discussion \& Future Work}
RoPE~\cite{su2024roformer} position embedding has ben widely adopted by both Large Language Models and Video Diffusion models. However, due to the intrinsic modeling difference, the end results are different. Our method will greatly suppress the sink-collapse phenomenon without retraining. However, retraining should further improve the generation quality. Due to the periodic nature or RoPE, we plan to further investigate its impact beyond sink-collapse. Alternative position embedding could be tested as well. Also, during training, blocking the sink tokens at local phase concentration maxima is also possible which can be computed following~\cref{eq:phase_kernel}. Although our method achieves infinite streaming generation, in order to further enhance the generation quality, we also plan to investigate models with larger size and integrating long-term memory into the model.

\section*{Limitations}
The primary objective of our work is to alleviate the sink-collapse phenomenon in ultra-long video generation frameworks such as LongLive~\cite{yang2025longlive} and Self-Forcing++~\cite{cui2025self}. Because our method extends these models into effectively infinite generation in a training-free manner, the overall generation quality is fundamentally bounded by the underlying LongLive and Self-Forcing++ architectures on which we build. First, the model has no mechanism for maintaining long-term memory. As a result, if an object leaves the frame and later reappears, or if it becomes occluded for an extended period, subject consistency may fail to hold. Second, the base model we rely on, a distilled 4-step Wan2.1-T2V-1.3B model, has limited representational capacity. When pushed to generate extremely long sequences for single prompts, such as videos lasting up to 12 hours, it can exhibit reduced visual diversity. As noted above, enhancing long-term memory and employing base models with enhanced capacity remain two important directions for our future work.

\end{document}